%% file: main.tex
\documentclass[runningheads]{llncs}
\usepackage{graphicx}
\usepackage{subcaption}

\usepackage{tikz}
\usepackage{comment}
\usepackage{amsmath,amssymb} 
\usepackage{color}
\usepackage{float}

\include{preamble}

\begin{document}
\pagestyle{headings}
\mainmatter
\def\ECCVSubNumber{4}  

\title{Semi-supervised Viewpoint Estimation with Geometry-aware Conditional Generation} 

\titlerunning{Semi-supervised Viewpoint Estimation with Conditional Generation}
%
\author{Octave Mariotti \and
Hakan Bilen\orcidID{0000-0002-6947-6918}}
\authorrunning{O. Mariotti, H. Bilen}
%
\institute{School of Informatics, University of Edinburgh, United Kingdom}
\maketitle

\begin{abstract}
There is a growing interest in developing computer vision methods that can learn from limited supervision.
In this paper, we consider the problem of learning to predict camera viewpoints, where obtaining ground-truth annotations are expensive and require special equipment, from a limited number of labeled images.
We propose a semi-supervised viewpoint estimation method that can learn to infer viewpoint information from unlabeled image pairs, where two images differ by a viewpoint change.
In particular our method learns to synthesize the second image by combining the appearance from the first one and viewpoint from the second one.
We demonstrate that our method significantly improves the supervised techniques, especially in the low-label regime and outperforms the state-of-the-art semi-supervised methods.
\end{abstract}

\section{Introduction}
\label{sec:intro}
\input{intro.tex}

\section{Related work}
\label{sec:relwork}
\input{relwork}

\section{Method}
\label{sec:method}
\input{method}

\section{Experiments}
\label{sec:experiments}
\input{experiments}

\section{Conclusion}
\label{sec:conclusion}
\input{conclusion}

~~~\\
\paragraph{Acknowledgements.} The authors acknowledge the support of Toyota Motor Europe.

\clearpage

%
%

\input{refs.bbl}
\end{document}

%% file: intro.tex

Large-scaled labeled datasets have been an important driving force in the advancement of the state-of-the-art in computer vision tasks.
However, annotating data is expensive and is not scalable to a growing body of complex visual concepts.
This paper focuses on the problem of viewpoint (azimuth, elevation, and in-plane-rotation) estimation of rigid objects relative to the camera from limited supervision where obtaining annotations typically involves specialized hardware and in controlled environments (\eg \cite{georgakis2016multiview,hodan2017t}) or a tedious process of manually aligning 3D CAD models with real-world objects (\eg \cite{xiang2014beyond}).
State-of-the-art viewpoint estimation methods~\cite{grabner20183d,zhou2018starmap,liao2019spherical} have shown to yield excellent results in the presence of large-scale annotated datasets, yet it remains unclear how to leverage unlabeled images.

One way of reducing annotation cost for viewpoint estimation is to produce synthetic datasets by rendering images of 3D CAD models in different views~\cite{liebelt2010multi}.
While it is possible to generate a large amount of labeled synthetic data with rendering and simulator tools and learn viewpoint estimators on them, discrepancies between the synthetic and real world images would make their transfer challenging.
Su~\etal~\cite{su2015render} show that overlaying images rendered from large 3D model collections on top of real images results result in realistic training images and boosts the viewpoint performance when they are used in training.
Motivated by the fact that accurate 3D models and diverse background scenes are not always available, domain randomization~\cite{tobin2017domain} addresses the reality gap by simulating a wide range of environments at training time and assume that the variability of the simulation is significant enough to generalize the real world.

Another line of work~\cite{thewlis2017unsupervised,thewlis2017dense,jakab2018conditional,rhodin2018unsupervised} learns representations from large sets of unlabeled images by self-supervised training and transfers the knowledge to a supervised down-stream viewpoint and pose estimation.
Thewlis~\etal~\cite{thewlis2017unsupervised,thewlis2017dense} learn sparse and dense facial landmarks based on the principles of equivariance and distinctiveness.
Jakab~\etal~\cite{jakab2018conditional} learn landmark detectors by factorizing appearance and geometry in a conditional generation framework.
However, these techniques are limited to predicting 2D landmarks, small in-plane rotations, insufficient to learn 3D objects in case of significant pose variations and the learned representations (\ie landmarks) are not semantically meaningful.
Most related to ours, Rhodin~\etal~\cite{rhodin2018unsupervised} propose a geometry aware autoencoder that learns to translate a person image from one viewpoint to another in a multi-camera setup.
Unlike \cite{rhodin2018unsupervised} that requires the knowledge of the rotation between each camera pairs, our method jointly learns the rotations between \emph{unknown viewpoints} and conditionally generating images.
This difference enables learning of our method on images from arbitrary views and extends its use to single camera setups.

\begin{figure}[t]
    \includegraphics[width=\textwidth]{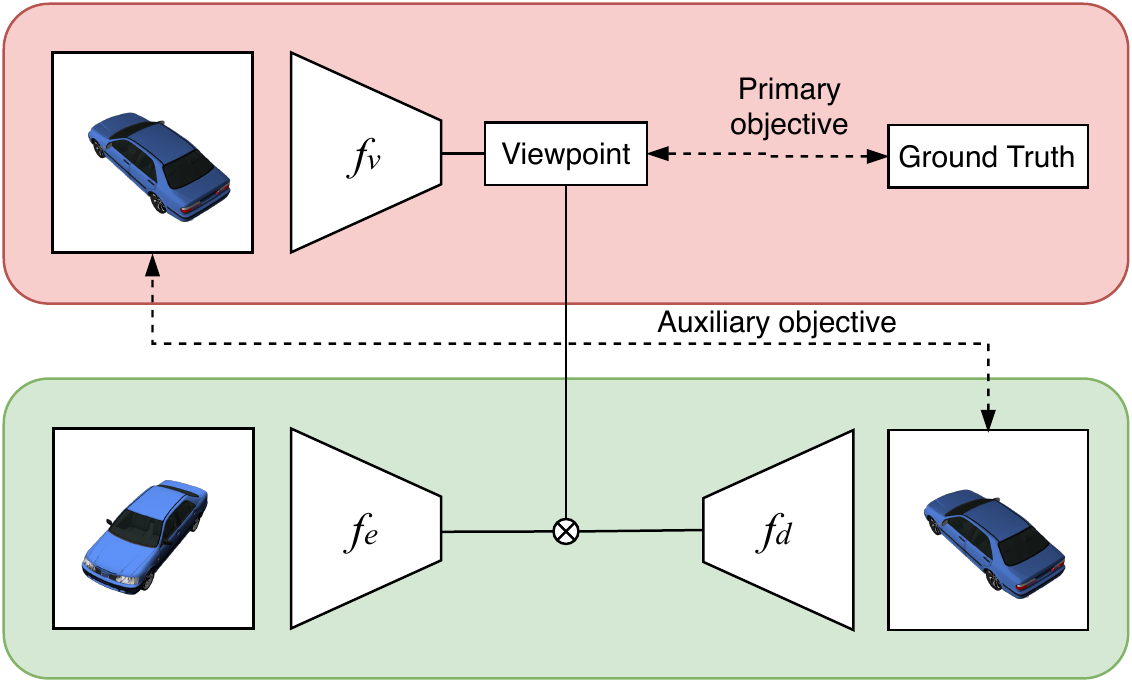}
    \caption{Overview of the semi-supervised framework. Our primary objective is to learn the camera viewpoint from the picture of an object. Given another picture of the same object, we also reconstruct the first using conditional generation to provide additional supervision.}
    \label{fig:global_arch}
\end{figure}

In this paper, we follow an orthogonal semi-supervised approach to the previous methods that focus on learning from synthetic data and self-supervision.
Semi-supervised learning is one of the standard approaches that learn not only from labeled samples but also unlabeled ones.
Many recent semi-supervised methods~\cite{tarvainen2017mean,berthelot2019mixmatch,xie2019unsupervised} employ a regularization techniques that encourages the models to produce consistent output distributions when their inputs are perturbed.
However, strong data augmentations (\eg MixUp~\cite{zhang2018mixup}) in \cite{berthelot2019mixmatch} that are used to perturb images in image classification are not applicable to viewpoint estimation, as mixing up two object images at different views produces an ambiguous input for viewpoint estimation.

Instead, we propose a semi-supervised method that is specifically designed for viewpoint estimation.
We pose this as the problem of image synthesis by conditioning on the viewpoint of objects in them.
In particular, our method takes in a pair of images that contain an object captured in different viewpoints, encodes the appearance of the object in the first image and estimates the viewpoint of the object in the second image, reconstruct the second image from them. An overview of our method is show on figure~\ref{fig:global_arch}.

While the conditional generation from image pairs forces the network to learn factorized representations for appearance and viewpoint and does not require label supervision, there is a high degree of ambiguity for representing the viewpoint in a deep neural network and no guarantee that the learned representation corresponds to ground-truth viewpoint.
Thus we address this challenge by simultaneously training the viewpoint estimator on a small set of labeled images to encourage the inferred viewpoints from unlabeled images to be consistent with the labeled ones.
We show that our method can effectively leverage the information in unlabeled images, improves viewpoint estimation with limited supervision and outperforms the state-of-the-art semi-supervised methods in a standard viewpoint estimation benchmark.

%% file: relwork.tex

\paragraph{Early supervised pose estimation.}
The early models proposed in object pose estimation use classical computer vision techniques, rely on matching image features like gradients or surface normals with predefined templates, either recovered from the object itself in a controlled setup, or by using 3D CAD models to obtain rough estimates \cite{hinterstoisser2011multimodal,hinterstoisser2012model}. These methods require pose supervision and have limited applicability due to their lack of generalization.

\paragraph{Recent supervised pose estimation.}
More recent methods typically split the pose estimation into two subtasks, object localisation and rotation estimation, use a CNN for each. 
Localisation is most often performed using pretrained models (\eg r-CNN\cite{girshick2015fast}) while rotation are recovered either by 3D bounding cube regression \cite{rad2017bb8,tekin2018real,grabner20183d} or viewpoint classification \cite{kehl2017ssd}. 
As excellent performances have been reported on controlled setups, focus has shifted towards specifications of the task, such as avoiding errors caused by symmetries and lowering the necessary supervision by removing depth maps \cite{rad2017bb8}, or targeting real-time performances \cite{kehl2017ssd}. Direct regression over the rotation space has also been explored and proved to be on par with other ways of recovering rotations \cite{mahendran20173d,liao2019spherical}, while multitask approaches have also been shown to help by adding keypoint detection for instance \cite{tulsiani2015viewpoints,zhou2018starmap}. Viewpoint estimation specific architectures that go beyond generic CNNs are also getting proposed, in a effort to tailor neural network to the characteristics of the task\cite{joung2020cylindrical,cohen2018spherical,esteves20173d,esteves2018cross}.

\paragraph{Pose estimation from synthetic data.}
As training labels on real images are expensive to obtain, many works try to reduce the cost of supervision. The most common approach is to use synthetic data obtained from 3D CAD models \cite{sundermeyer2018implicit,tan2018indirect,su2015render}, however, these methods tend to have issues generalizing to real data. Other works explore using extremely limited sets of poses \cite{rhodin2018unsupervised,kanezaki2018rotationnet}, mainly because of dataset restrictions.

\paragraph{3D reconstruction.}
Another line of work focuses on producing a 3D reconstruction of the object from 2D views by geometry-aware deep representations. However, as they are only interested in the 3D-aware representation, they tend to consider pose information as an already-acquired supervision. Nonetheless, several recent works show that pose supervision is not strictly required to produce a 3D model, either voxel-based \cite{yan2016perspective,yang2018learning,tulsiani2018multi}, point clouds \cite{insafutdinov2018unsupervised}, or 3D meshes \cite{kato2018neural}. The mesh approach was also extended in an unsupervised way \cite{kanazawa2018learning}. In this case, pose is learned jointly with the reconstruction and supervision is done by projection the 3D shape in camera space. The 2D image obtained is then compared with the ground-truth segmentation mask. These works involve heavy networks to deal with full 3d representations, and a complex differentiable rendering stage. In contrast, we aim for a fully convolutional, more flexible architecture.

\paragraph{Geometry-aware representations.}
Another related line of approaches involves producing lighter weight representations that describe the geometry of the object while being sensitive to pose. Often, these are designed following an equivariance principle, that is, applying a transformation \eg a rotation to the object will have the effect of transforming the representation in similar way. Precursory works specifically targeting equivariance rely on autoencoding architecture and constrain the encoding to respect the structure of the data \cite{kulkarni2015deep}. A more involved approach consists in entangling a learned embedding with a rotation. This has first been proposed on feature maps and 2D rotations \cite{jaderberg2015spatial}, then adapted to general representations \cite{worrall2017interpretable} and applied on full 3D rotations \cite{rhodin2018unsupervised}, albeit with a very restricted set of poses.
Other rotation-specific equivariant representations were also designed by adapting CNNs to operate on spherical signals \cite{cohen2018spherical,esteves20173d,esteves2018cross}. These spherical CNNs rely on heavy 3D supervision and typically operate on a coarse scale due to their use of Fourier transform, but their construction guarantees good results on rotation estimation.

\paragraph{Keypoint-based methods.}
Keypoints are a natural equivariant representation: they describe the pose and it is intuitively possible to discover them without supervision. 2D keypoints have been discovered on humans and faces with \cite{jakab2018conditional} or without \cite{thewlis2017unsupervised} reconstruction. 3D keypoints are used in case of full 3D rotations \cite{suwajanakorn2018discovery}, however, no approach has been shown to reliably estimate them without strong pose supervision. Mapping the image pixels to a sphere has also been explored as a continuous generalization of keypoints, but this technique faces the same issues as its discrete counterpart \cite{thewlis2017dense,thewlis2018modelling}.

\paragraph{Generative-based methods.}
Recent advances in generative adversarial networks have allowed frameworks to learn geometry-aware representations, through the generation of images under different viewpoint \cite{nguyen2019hologan,mustikovela2020self}.These methods are still experimental and are still subject to a certain degree of unsuitability, but show a promising and novel angle of attack on viewpoint estimation.

%% file: method.tex

\subsection{Supervised viewpoint estimation}
Assume that we are given a set of $m$ labeled images with their ground-truth viewpoints, \wrt a fixed camera, $\mathcal{T}=\{(I_i,v_i)\}_{i=1}^m$, where $I\in\mathcal{I}$ is an RGB image and $v=(a^1,a^2,a^3)\in\mathcal{V}$ is a 3-dimensional vector represented in azimuth, elevation and in-plane rotations respectively, their values are between $0$ and $2\pi$.
There exists several ways to represent $v$, most common methods being the axis-angle representation, a unit quaternion or a rotation matrix. To simplify the learning procedure, we model $v$ simply by a three-dimensional vector interpreted as the camera position in object-centric coordinates. 
We wish to learn a mapping from an image to its viewpoint $f_v~:~\mathcal{I}\rightarrow\mathcal{V}$ such that $f_v(I;\theta_v)=v$ where $\theta_v$ are the parameters of $f_v$.
One can learn such a mapping by minimizing the following empirical loss over the set $\mathcal{T}$ \wrt $\theta_v$:
\begin{equation}
	\sum_{(I,v)\in\mathcal{T}} \lvert\lvert f_v(I;\theta_v) - v \rvert\rvert^2.
	\label{eq:optpose}
\end{equation}

\subsection{Geometry-aware representation}
We are also given a set of $n$ unlabelled image pairs $\mathcal{U}=\{(I_i,I'_i)\}$ where each pair contains two images of an object instance (\eg airplane, car, chair) that are captured at two different viewpoints.
We assume that the ground-truth viewpoints of the images are not available and we wish to improve the performance of our viewpoint predictor $f$ by leveraging the information in the unlabeled images.

A commonly used tool for unsupervised learning is autoencoder that encodes its input $I$ into a low dimensional encoding $f_e(I;\theta_e)$ via an encoder network $E$ and maps the encoding to the input space, \ie $f_d(f_e(I;\theta_e);\theta_d)$, via a decoder network $f_d$ to reconstruct the input.
The encoder and decoder are parameterized by $\theta_e$ and $\theta_d$ respectively.
Although autoencoders can successfully be utilized to learn informative representations that can reconstruct the original image, there is no guarantee for the embeddings to encode the 3D viewpoints of objects in a disentangled manner.

One solution to relate an embedding of an object in image $I$ to its viewpoint $v$ involves a conditional image generation technique.This was first proposed in \cite{worrall2017interpretable} for in-plane rotations and extended in \cite{rhodin2018unsupervised} for 3D ones, 
In particular, given an image pair $I$ and $I'$ that contain the same object viewed from two different points and also given the viewpoint of from which the object is seen in the images, this method couples the viewpoint and the appearance of the object in the encoding.
To this end, the embedding of image $I$, $f_e(I)$ is transformed by using the rotation $R(v')$ where $v'$ is the viewpoint in image $I'$ and $R(v') \in SO(3)$ computes the rotation matrix associated to $v'$.
The rotated embedding is then decoded, \ie $f_d(R(v') \times f_e(I;\theta_e));\theta_d)$, to reconstruct not the input $I$ but $I'$ by minimizing the following loss \wrt the parameters of the encoder and decoder:
\begin{equation}
	\sum_{(I,I',v')\in\mathcal{U}}\lvert\lvert f_d(R(v')\times f_e(I;\theta_e));\theta_d) - I' \rvert\rvert^2
	\label{eq:optgeom}
\end{equation} where the output of the encoder $f_e(I;\theta_e)$ is designed to be $3\times k$ dimensional such that it can be rotated by the rotation matrix $R(v')$.

This presents a slight variation over the framework in \cite{rhodin2018unsupervised} as the rotation here is absolute instead of relative. This means that the embedding $f_e(I;\theta_e)$ should represent the object from an canonical viewpoint instead of the one from which it appears in $I$.

This formulation enables the method to learn a ``geometry aware'' representation that can relate the viewpoint difference in 3D space to its projection in pixel space.
However, it requires the ground-truth viewpoint for each image $I'$, which limits the applicability of the method to supervision-rich setups.
To address this limitation and extend learning of the geometry-aware representations to image pairs with unknown viewpoints, we propose an analysis by synthesis method.
To this end, we predict the viewpoint as $\hat{v}=f_v(I;\theta_v)$ for $I$ by using the viewpoint estimator $f$, and substitute it with $R(v')$ in \cref{eq:optgeom}:
\begin{equation}
	\sum_{(I,I')\in\mathcal{U}}\lvert\lvert f_d(R(f_v(I';\theta_v))\times f_e(I;\theta_e));\theta_d) - I' \rvert\rvert^2
	\label{eq:optgeom2}
\end{equation} 

This formulation models the reconstruction loss as a function of viewpoint predictor $f$ and therefore allows the gradients to flow in the pose regression network without any viewpoint supervision. Furthermore, working with absolute viewpoints not only allows a more straightforward optimization as we only need one viewpoint estimation whereas two would be needed to compute a relative pose, it also makes learning an encoding easier as it factors out the burden of estimating the pose.

\subsection{Semi-supervised viewpoint prediction}
Our hypothesis is that a successful reconstruction of $I'$ requires an accurate viewpoint estimation.
However, given high-capacity encoder and decoder architectures, accurate viewpoints enable high-fidelity reconstructions, the converse is not necessarily true as the viewpoints in the encoding can be represented in infinite different ways and there is no guarantee that the learned viewpoints for the images will match with their ground-truth view. 
For instance, the output of the viewpoint estimator can be distributed between $0$ and $\pi$ for each angle instead of the entire range of $[0, 2\pi)$ or the angles can be mapped to a non-linear and uninterpretable space, while the network preserves its reconstruction performance.
Thus, we propose a semi-supervised formulation in which the estimated viewpoints are regularized as below by optimizing the combined loss terms in \cref{eq:optpose} and \cref{eq:optgeom2}:
\begin{equation}
	\min_{\theta_v,\theta_e,\theta_d} \sum_{(I,v)\in\mathcal{T}}  \lvert\lvert f_v(I;\theta_v) - v \rvert\rvert^2 + \lambda \sum_{(I,I')\in\mathcal{T}\cup\mathcal{U}}\lvert\lvert f_d(R(f_v(I';\theta_v))\times f_e(I;\theta_e));\theta_d) - I' \rvert\rvert^2
	\label{eq:combined}
\end{equation} where $\lambda$ is a tradeoff hyperparameter between the supervised and unsupervised loss terms.
In words, the formulation allows gradients for the unsupervised loss to flow in the viewpoint network $f_v$, and the supervision imposed on the viewpoint space in turn constrains the learned representation to capture the structure of the object.

The supervision provided by the reconstruction task brings the question of unsupervised viewpoint estimation using no pose labels. While theoretically possible, we find that it is likely to fail in complex scenarios, as the supervision signal is too weak to provide good viewpoint supervision. In particular, symmetries in real world objects push the learned pose towards degenerates solutions. This is further demonstrated in sections~\ref{pred} and~\ref{unsup}

%% file: experiments.tex

\subsection{Dataset}
We use the popular Shapenet \cite{chang2015shapenet} dataset that consists of a large bank of 3D CAD models, classified in different object categories. This makes obtaining a large number of views spreading various viewpoints fairly straightforward, as well as acquiring several views of the same object, a feature often absent in other 3D datasets like Pascal3D \cite{xiang2014beyond}. Because we render the 3D models, we automatically know the ground truth viewpoint as well, making data labeling a triviality. We mainly focus on three object categories, aeroplanes, cars and chairs, as they offer enough models to build a diversified image dataset. For each category, we render each model with 10 randomly selected viewpoints, with azimuth ranging the complete $360^o$ rotation and elevation selected from $-20^o$ to $40^o$. The final datasets contain 40460, 36760 and 67790 images for the aeroplane, car and chair category respectively. We split the data in training, validation and testing sets, accounting for 70, 10 and 20 percent of the whole dataset respectively. To simulate a semi-supervised setup, we further split the training set by randomly selecting a subset of the data to act as the labeled set, the rest acting as unlabeled. We adjust the ratio of labeled samples in our experiments to show the effect of varying degrees of supervision. The splits are made on a model basis, that is, the different views from the same 3D model are either all labeled or all unlabeled.

To evaluate our framework, we use two popular metrics in viewpoint estimation \cite{tulsiani2015viewpoints,tulsiani2018multi,insafutdinov2018unsupervised}, the accuracy at $30^o$, and the median angular error in degrees. The accuracy is computed as the ratio of predictions within $30^o$ of the ground truth viewpoint and gives a rough estimate of the network performances. The aggregator for angular error is chosen to be the median rather than the mean as it is less biased by outliers which are common in pose estimation due to symmetries.

\subsection{Implementation details}

\begin{figure}[t]
	\centering
	\includegraphics[width=\linewidth]{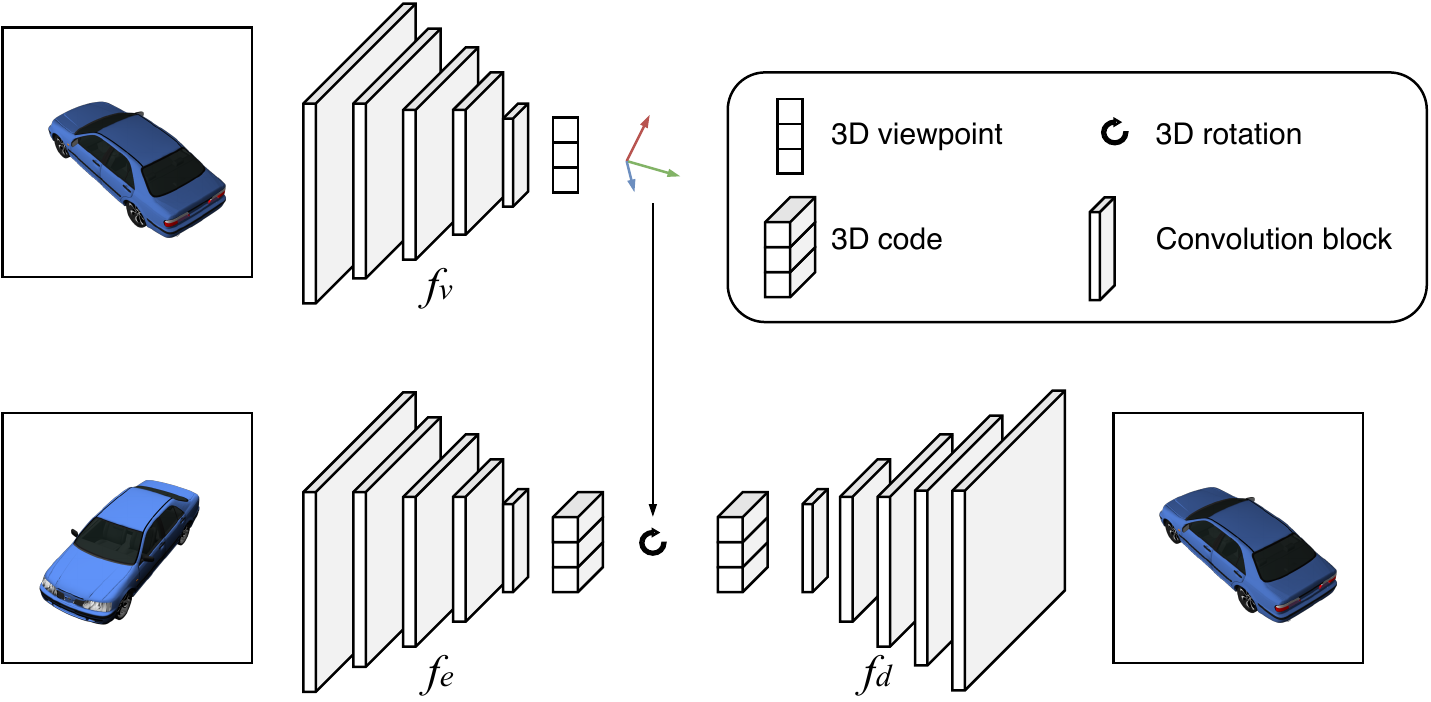}
	\caption{Detailed architecture of the network. The viewpoint estimator $f_v$ outputs the camera coordinates. This prediction is transformed in a rotation matrix, which is used to rotate the code produced by the encoder $f_e$. This rotated embedding is given to the decoder $f_d$ to reconstruct the original image.}
	\label{fig:model}
\end{figure}

We model $f_e$, $f_d$ and $f_v$ with convolutional neural networks. We use a simple design, stacking several convolutional blocks with batch normalization and ReLU activation function. The encoder network has five blocks each consisting of two convolutions layers, with the second of each block using a stride of 2 in order to reduce the spatial dimension. All layers use 3 by 3 convolutions with channel count starting at 32 and doubling each block. On top of this, we use a fully connected layer to obtain the embedding. In order to interpret it as a geometric representation, we group the embedding values by triplets, effectively creating a collection of points in 3D space. This representation can then be rotated using the viewpoint rotation matrix. The architecture of $f_d$ is simply a mirrored version of that of $f_e$. A schematic version of our framework is presented in figure~\ref{fig:model}.

For the reconstruction objective, we use perceptual loss \cite{johnson2016perceptual}, as it provides supervision of higher quality, translating in better learning signals for the viewpoint estimation. All training is done with the ADAM optimizer \cite{kingma2014adam} with default parameters and a batch size of 64. To prevent potential overfitting caused by the reconstruction task, we use early stopping, halting the training when no improvements are observed on the validation set for 30 epochs.
We set the hyperparameter $\lambda$ to be equal to the ratio between labeled and unlabeled samples. This way, when summed over the whole sets, the contributions of both losses are evened out.

\subsection{Viewpoint estimation}

\begin{table}[t]
  \centering
  \begin{tabular}{lccccccc}
    \toprule
    \multirow{2}{*}{Method} & \multirow{2}{*}{Labels  (\%)} & \multicolumn{2}{c}{aeroplane} & \multicolumn{2}{c}{car} & \multicolumn{2}{c}{chair}\\
                  &     & Acc   & Err   & Acc   & Err   & Acc   & Err\\
   \midrule
    Regression    & 100 & \B 87.3 &  6.9 & 89.3 &  6.2 & 88.9 &  8.4 \\
    Ours          & 100 & \B 87.3 &  \B 6.1 & \B 91.4 &  \B 4.6 & \B 89.7 &  \B 7.8 \\
    \midrule
    Regression    &  25 & 80.7 &  8.9 & 79.4 &  9.8 & 80.8 & 12.2 \\
    Ours          &  25 & \B 84.9 &  \B 6.4 & \B 86.6 &  \B 5.8 & \B 86.2 &  \B 8.5 \\
    \midrule
    Regression    &  10 & 75.6 & 12.1 & 72.3 & 13.1 & 71.8 & 16.5 \\
    Ours          &  10 & \B 83.2 & \B  6.5 & \B 83.7 & \B  6.4 & \B 81.0 & \B  9.4 \\
    \midrule
    Regression    &   5 & 70.4 & 15.1 & 65.9 & 17.7 & 68.4 & 19.2 \\
    Ours          &   5 & \B 81.4 & \B  7.4 & \B 73.8 & \B  9.0 & \B 76.3 & \B 15.1 \\
    \midrule
    Regression    &   1 & 54.2 & 29.5 & 45.1 & 36.3 & 59.1 & 28.6 \\
    Ours          &   1 & \B 64.9 & \B 17.1 & \B 62.4 & \B 14.5 & \B 57.9 & \B 25.1 \\
  \bottomrule
  \end{tabular}
  \caption{Viewpoint prediction in terms of accuracy and error rates for varying label supervision. Regression denotes a supervised trained network trained on the corresponding proportion of labeled data.}
  \label{tab:perfs10}
\end{table}

We compare the results of our method with a simple regression baseline, as well as Mean Teacher, a state-of-the-art semi-supervised approach. Though it was proposed for classification, it is a generic approach that can therefore be extended to viewpoint regression. Training is done using 10 views per model, with varying degrees of supervision. The baseline is simply set as a viewpoint estimator without any added secondary objective, in order to study the effect adding reconstruction to the framework has.

The quantitative results in \cref{tab:perfs10} show that our method outperforms simple regression in all cases. Unsurprisingly, performances are directly correlated with the amount of labeled data for all methods. It is worth noting that even when using 100\% of the labels, our method still outperforms simple regression, showing that simply adding a reconstruction task helps refine the network predictions. However, the gap in performances increases more when lowering supervision, as the regression task is losing training samples while ours can still leverage them in a self-supervised way, demonstrating the effectiveness of reconstruction as a proxy for viewpoint estimation. When training with very low supervision, prediction tend to drop sharply, as symmetries in the object make viewpoint estimation too difficult, and the reconstruction task becomes less effective. Indeed, producing an image from a symmetric viewpoint still provides decent minimization of the reconstruction loss. A significant failure of our system can be observed when using only 1\% of the labels on the chair category, which accounts to only 470 labeled images. Further details are discussed in section~\ref{pred}.

\begin{table}[t]
  \centering
  \begin{tabular}{lccccccc}
    \toprule
    \multirow{2}{*}{Method} & \multirow{2}{*}{Labels (\%)} & \multicolumn{2}{c}{aeroplane} & \multicolumn{2}{c}{car} & \multicolumn{2}{c}{chair}\\
                  &     & Acc   & Err   & Acc   & Err   & Acc   & Err\\
   \midrule
   Mean teacher \cite{tarvainen2017mean}  &  10 & 81.4 & 10.3 & 72.4 & 13.8 & 68.9 & 19.0\\ 
   Ours                                   &  10 & \B 83.2 & \B  6.5 & \B 83.7 & \B  6.4 & \B 81.0 & \B  9.4 \\
    
    \midrule
    Mean teacher \cite{tarvainen2017mean}  &   1 & 28.9 & 44.0 &  8.5 & 67.7 & 34.3 & 39.5\\
    Ours                                   &   1 & \B 64.9 & \B 17.1 & \B 62.4 & \B 14.5 & \B 57.9 & \B 25.1 \\    
    
  \bottomrule
  \end{tabular}
  \caption{Comparison to the Mean Teacher~\cite{tarvainen2017mean}\\ in terms of viewpoint accuracy and error rate. }
  \label{tab:comp2mt}
\end{table}

We are also able to outperform mean teacher~\cite{tarvainen2017mean}, demonstrating how building a problem-specific approach can easily lead to better performances (table~\ref{tab:comp2mt}). Mean teacher relies on prediction consistency over the unlabeled set, using averaged models to predict soft targets. This constrains the learning procedure to be stable during training, making predictions more reliable. However, reliably wrong predictions will not be detected, in which case the unlabeled set is of no help. This is a common pitfall in viewpoint estimation because of the symmetries. In contrast, our method always provides a supervision signal in case of wrong reconstruction, effectively alleviating the issue. Similarly to our approach, mean teacher tends to fail when supervision is scarce, as illustrated by its results with 1\% supervision.

\subsection{Prediction analysis}
\label{pred}
\begin{figure}[!ht]
    \centering
	 \begin{subfigure}[b]{0.45\textwidth}
        \includegraphics[width=\textwidth]{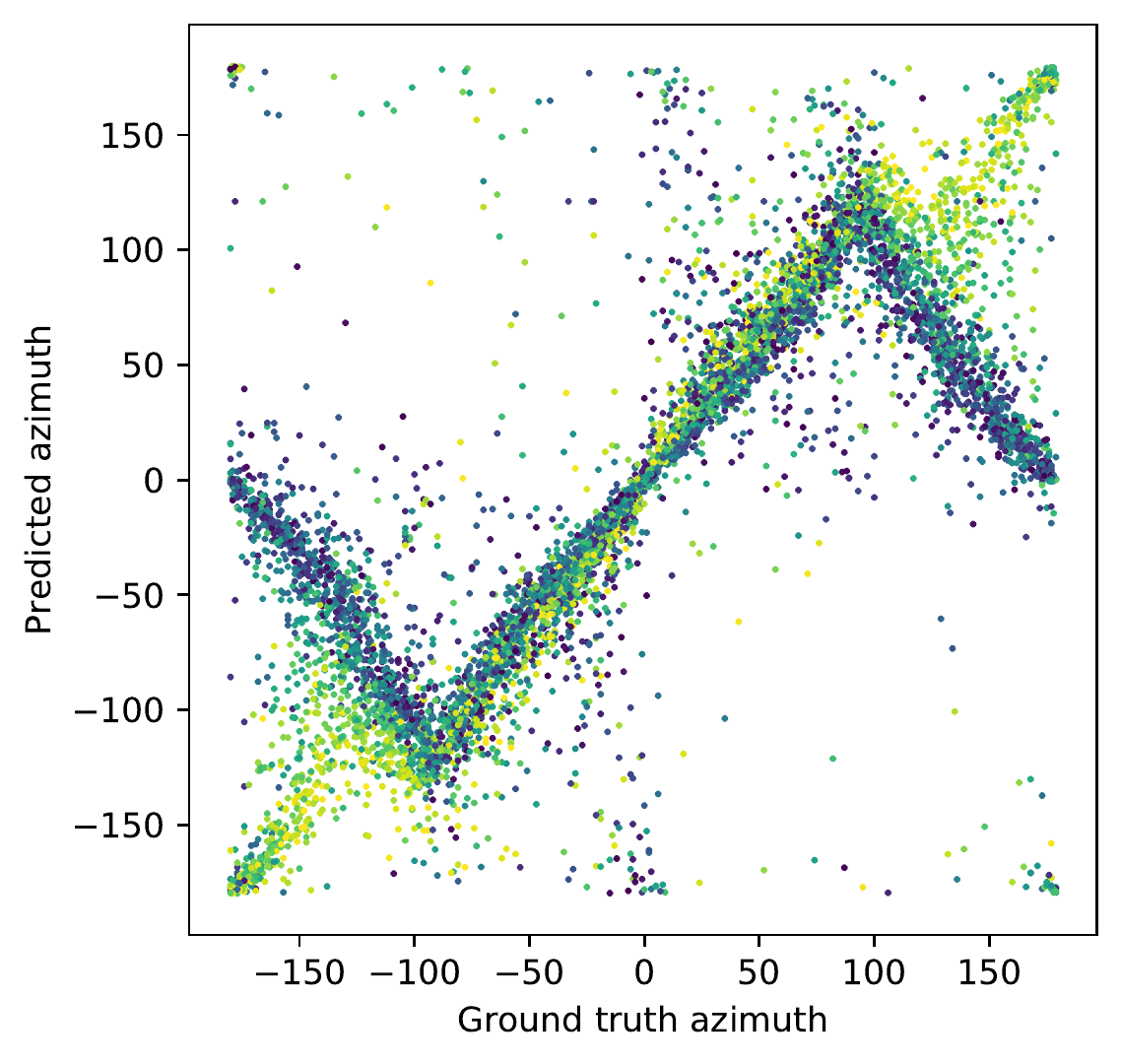}
        \caption{Ours, 1\% labels}
        \label{fig:our_1percent}
    \end{subfigure}
    \begin{subfigure}[b]{0.45\textwidth}
        \includegraphics[width=\textwidth]{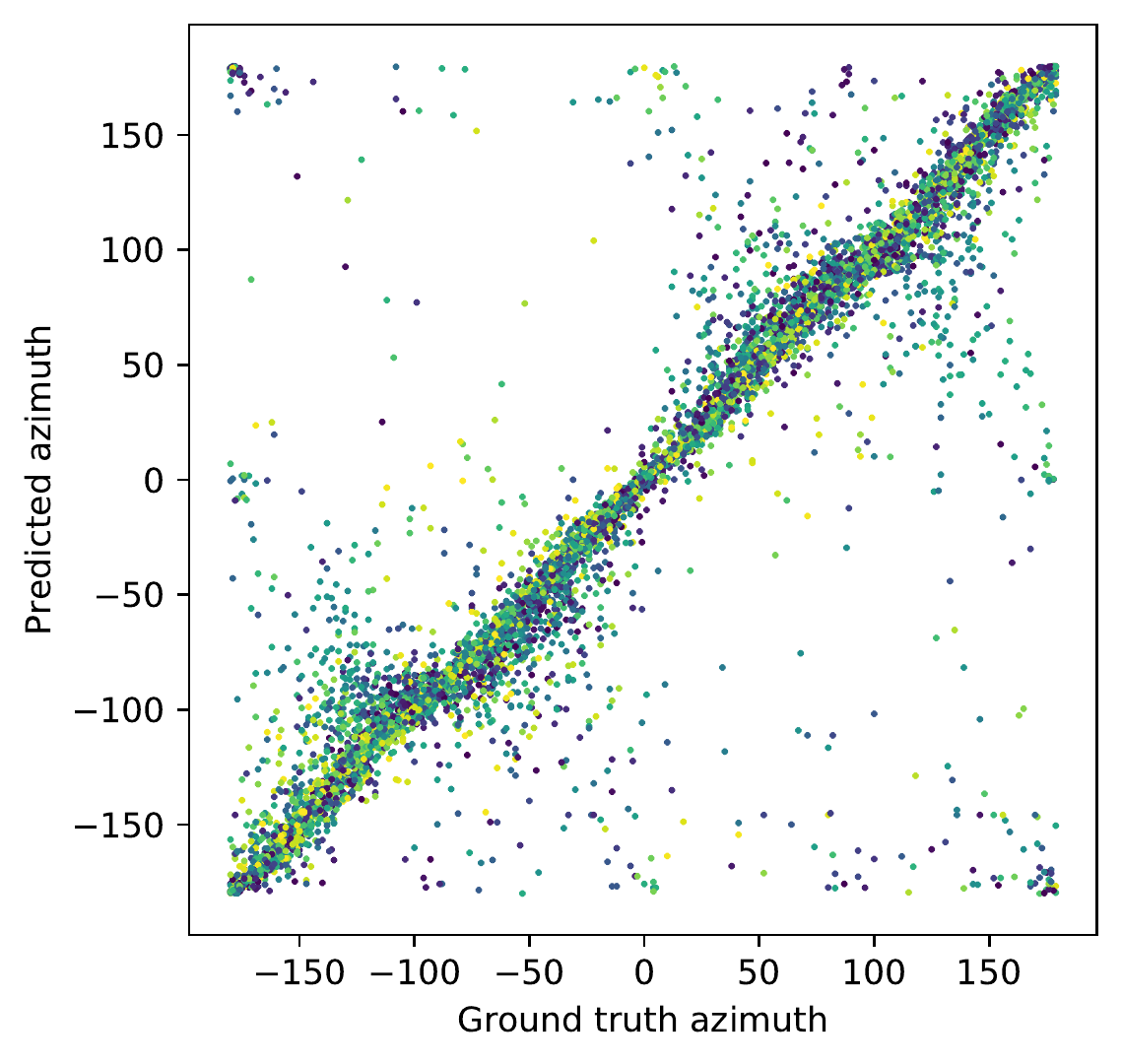}
        \caption{Ours, 10\% labels}
        \label{fig:our_10percent}
    \end{subfigure}
    \caption{Predicted vs ground truth azimuth for our method on test samples.\\Each point is colored with ground truth elevation.}
	\label{fig:our_sucfail}
\end{figure}

\begin{figure}[!ht]
    \centering
	 \begin{subfigure}[b]{0.45\textwidth}
        \includegraphics[width=\textwidth]{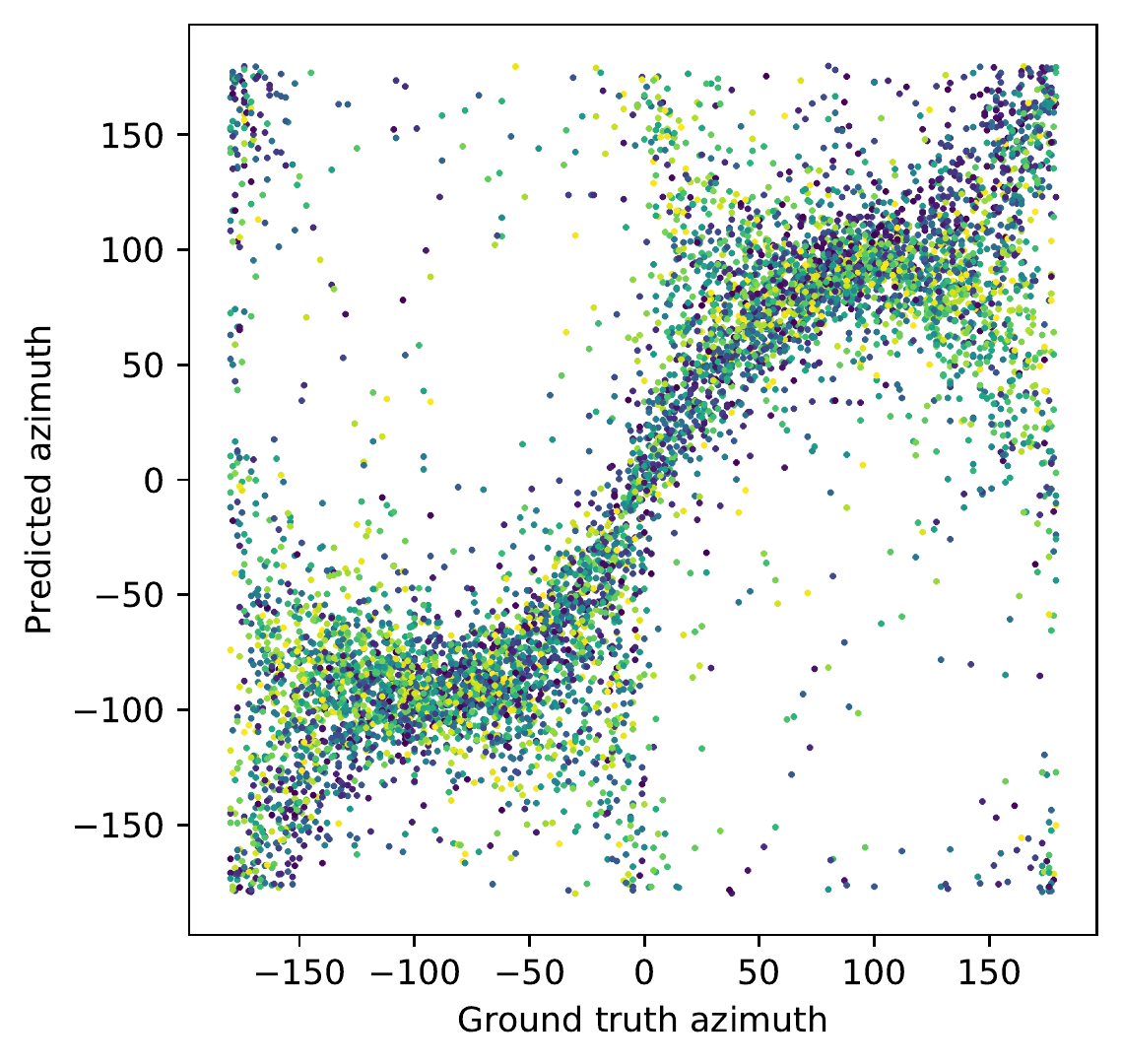}
        \caption{Regression only, 1\% labels}
        \label{fig:1percent}
    \end{subfigure}
    \begin{subfigure}[b]{0.45\textwidth}
        \includegraphics[width=\textwidth]{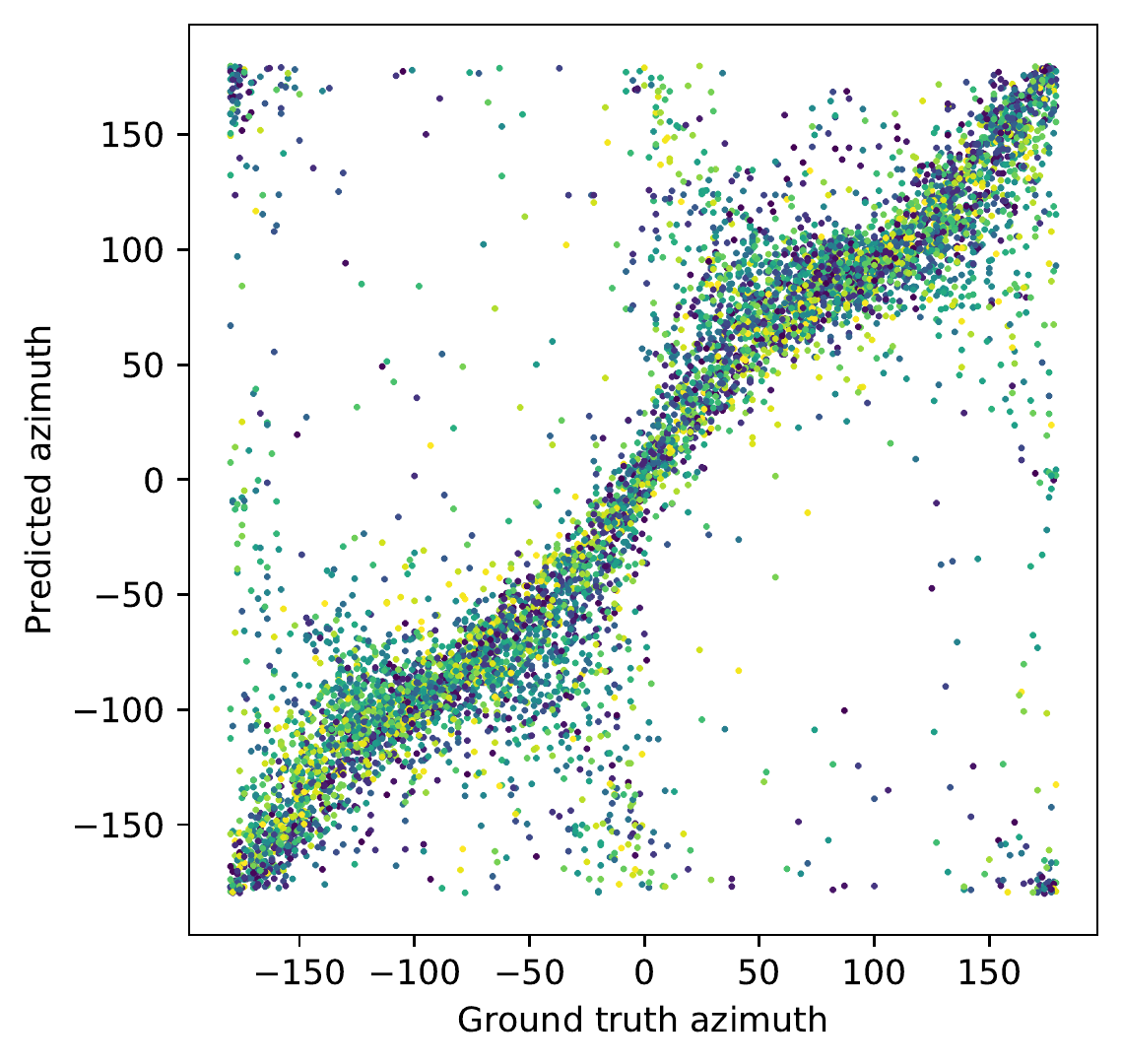}
        \caption{Regression only, 10\% labels}
        \label{fig:10percent}
    \end{subfigure}
    \caption{Predicted vs ground truth azimuth for simple regression on test samples.\\Each point is colored with ground truth elevation.}
	\label{fig:sucfail}
\end{figure}

An interesting phenomenon can occur when the supervision is low : the symmetries of the object will cause the emergence of degenerate solutions. If we consider a pair of images from two symmetric viewpoints, not only is it easy to mistake one viewpoint for the other when trying to learn it, reconstructing the wrong image is also not very penalizing. Those effects combined can push the network in a local minimum from which escaping becomes impossible, as the reconstruction objective is likely to push the viewpoint estimation back. Figure~\ref{fig:our_1percent} shows this behavior with chairs, as 1\% supervision sees the predicted azimuth ping back and forth when it should complete a full rotation. Increasing the supervision solves this issue, though we can still spot the occasional mistake (figure~\ref{fig:our_10percent}).

We also compare with predictions of a simple regression. We can see on figure~\ref{fig:sucfail} that while the global structure of predictions are similar, a simple regression involves more noise in the labels. In contrast, the predictions from our method are much finer, as the additional reconstruction provided gradients to correct small mistakes and give confidence to the viewpoint estimator.

\subsection{Multiview supervision}

\begin{table}[t]
  \centering
  \begin{tabular}{lccccccc}
    \toprule
    \multirow{2}{*}{Views} & \multicolumn{2}{c}{aeroplane} & \multicolumn{2}{c}{car~~~} & \multicolumn{2}{c}{chair}\\
       & Acc   & Err   & Acc   & Err   & Acc   & Err\\
       \midrule
     2 & 56.9 & 25.1 & 40.7 & 39.1 & 30.1 & 44.7 \\
     5 & \B 59.1 & \B 23.4 & 48.0 & 32.6 & \B 49.0 & \B 30.5 \\
    10 & 34.4 & 45.2 & \B 54.4 & \B 26.1 & 26.7 & 49.2 \\

  \bottomrule
  \end{tabular}
  \caption{Viewpoint prediction performance for varying number of views,\\ performed at 10\% of the labels.}
  \label{tab:numviews}
\end{table}

We conducted experiments with varying number of views per model to assess the importance of multi-view supervision. We compared the performances of a network trained on 2, 5, or 10 views per model. For a fair comparison, we made sure that the training set size was constant throughout the different experiments: we truncated the 5 and 10 views sets in order to match the size of the 2 views for each model. This means that models trained on those sets will see more of each model, but less models in total. Similarly, the viewpoint labels will be concentrated on the less models. Training was conducted with 10\% of the labels in all cases.

The results in table~\ref{tab:numviews} show that multi-view supervision seems to be profitable for the network, as increasing the number of views leads to increased performances. One way to interpret this result is that more views allow the encoder to build a representation more representative of the global structure of the object, therefore making the reconstruction supervision more effective. Indeed, it will be easier for the network to learn global information about the object when presented more views as the probability that the views cover the whole object increases, while finding correspondences has to be performed across different models when the view count is low.

However, because the number of labeled models also decreases, there is a risk that not enough will be available to learn a correct viewpoint estimator, harming performances as seen with the chairs and cars. The viewpoint estimator falls in this case in a local minimum, as depicted in section~\ref{pred}. We theorize then that multiple views benefit the framework, as long as it does not come to the detriment of variety in pose labels.

\subsection{Unsupervised viewpoint estimation}
\label{unsup}
\begin{figure}[b]
    \centering
    \includegraphics[width=.15\textwidth]{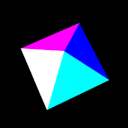}
    \includegraphics[width=.15\textwidth]{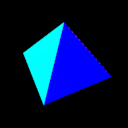}
    \includegraphics[width=.15\textwidth]{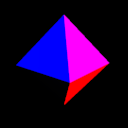}
    \includegraphics[width=.15\textwidth]{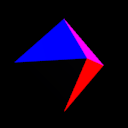}
    \includegraphics[width=.15\textwidth]{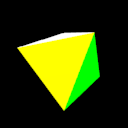}
    \includegraphics[width=.15\textwidth]{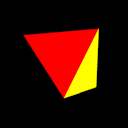}
    \caption{Example views from the toy dataset}
    \label{fig:Toy_sample}
\end{figure}

\begin{figure}[t]
    \centering
	 \begin{subfigure}[b]{0.45\textwidth}
        \includegraphics[width=\textwidth]{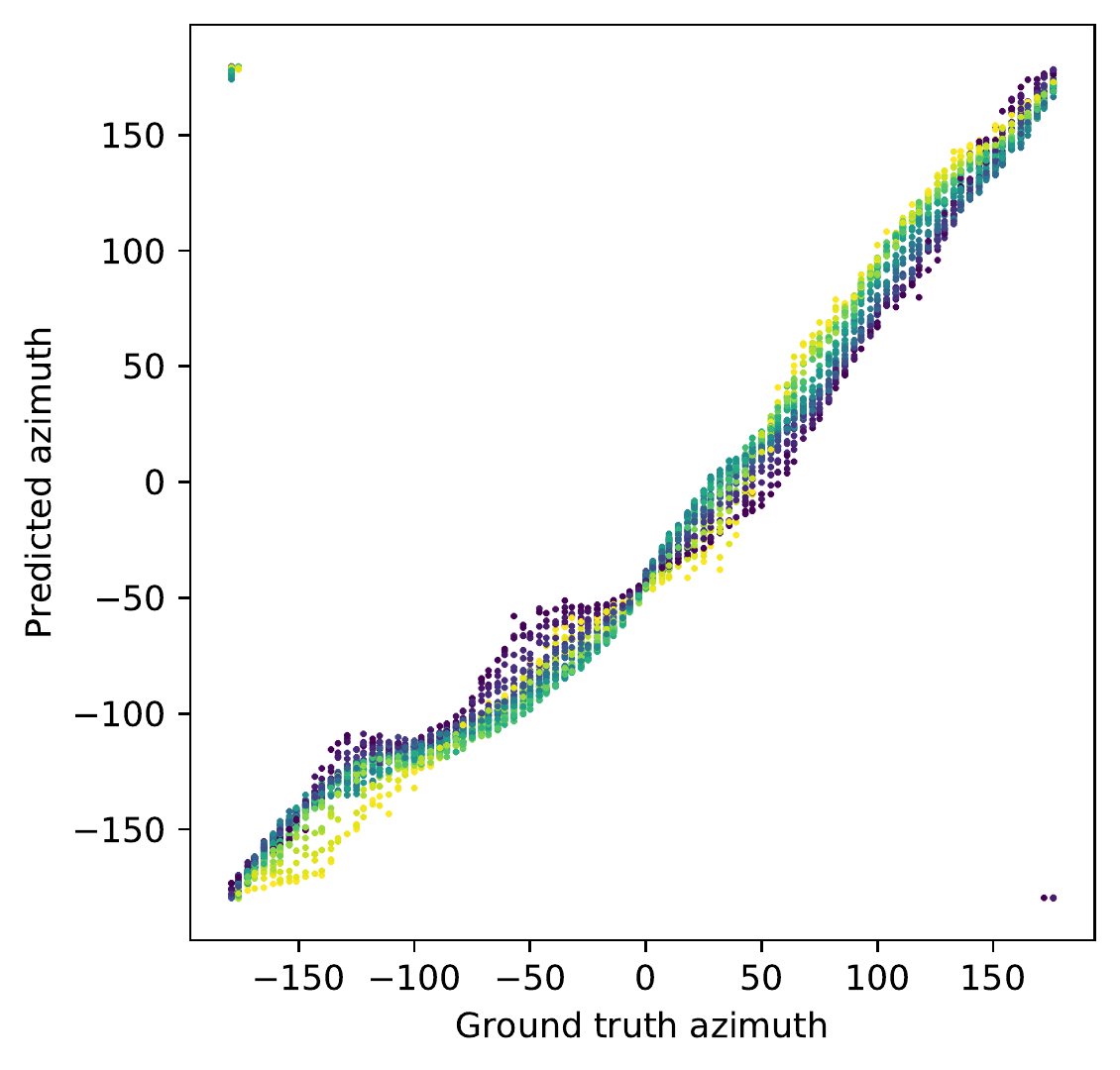}
        \caption{Toy data}
        \label{fig:unsup_toy}
    \end{subfigure}
    \begin{subfigure}[b]{0.45\textwidth}
        \includegraphics[width=\textwidth]{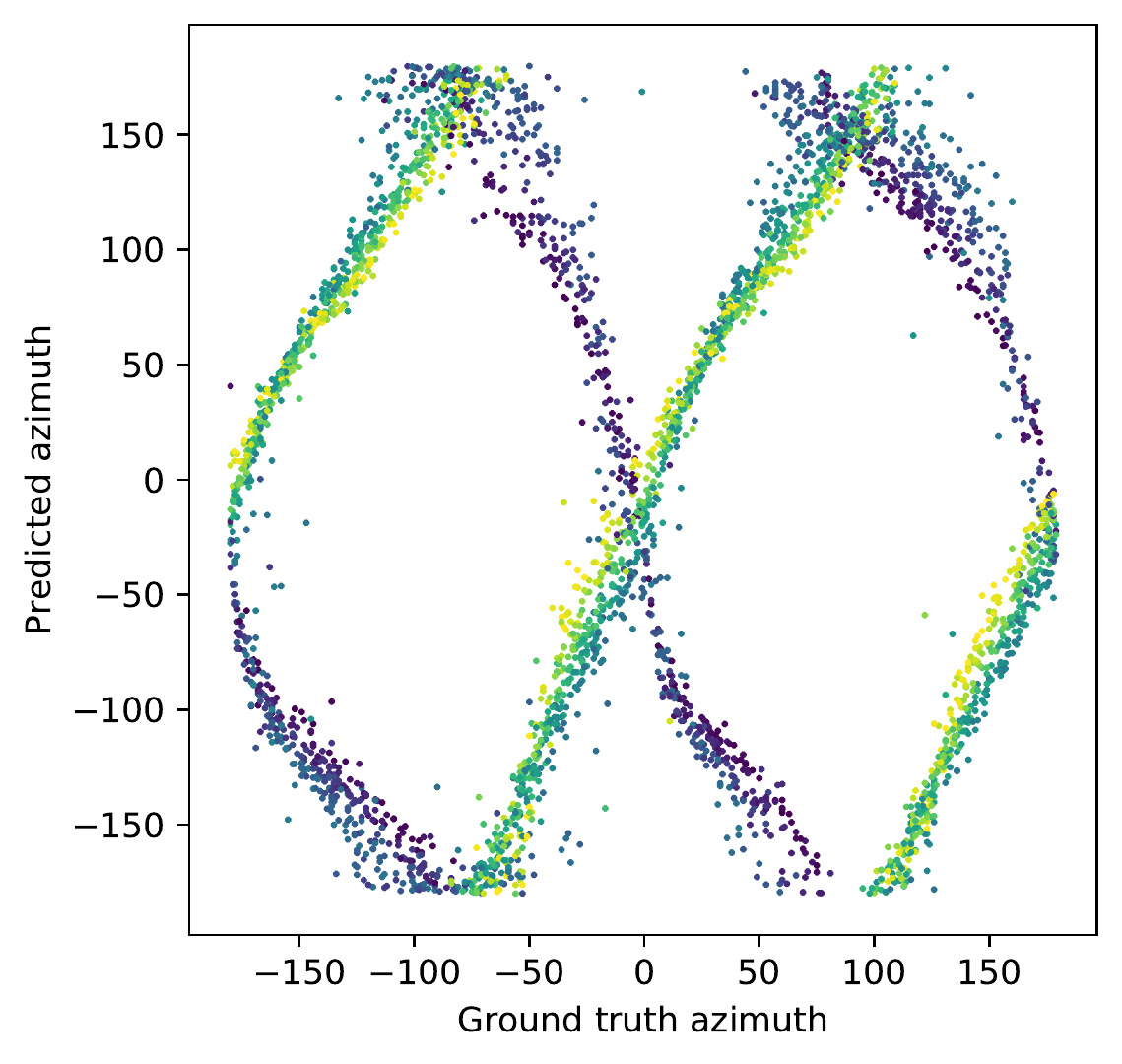}
        \caption{Cars}
        \label{fig:unsup_cars}
    \end{subfigure}
    \caption{Predicted vs ground truth azimuth without pose labels.\\Each point is colored with ground truth elevation.}
	\label{fig:unsup_pred}
\end{figure}

In these experiments, we assess the feasibility of training our framework in an unsupervised way, that is, without any pose labels, relying only on reconstruction. To this end, we designed a very simple dataset consisting of views from a single octahedron, with different colors on each face in order to break symmetries (figure~\ref{fig:Toy_sample}). The results of the viewpoint prediction shown on figure~\ref{fig:unsup_toy} confirm that we can indeed learn the correct structure of the pose in easy cases. Having no reference point, this is learned up to a random rotation, which we recovered using the validation set by minimizing the distance between ground truth and prediction.

However, we found that our model was unable to learn the correct pose when confronted with more complex data, \eg cars (figure~\ref{fig:unsup_cars}). We observe that the learned pose wraps twice around the pose space while the ground truth completes only one rotation. This is easily explainable as cars exhibit a strong symmetry when flipped $180^o$ around the vertical axis. The viewpoint predictor therefore identified these two poses to the same point. We also note that above horizontal views are treated differently from below ones. This is explained by the perceived way the object is rotating depending whether the observer is located above or below the object.

\subsection{Novel view synthesis}
\begin{figure}[t]
    \centering
    \includegraphics[width=\textwidth]{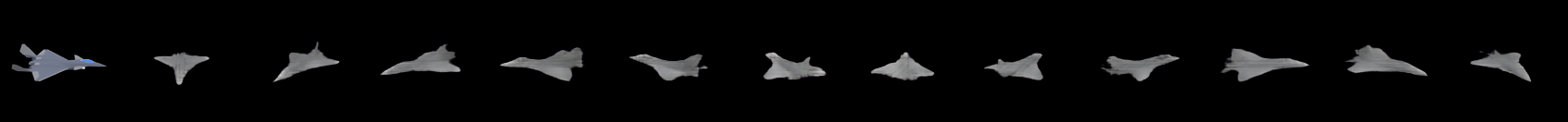}
    \includegraphics[width=\textwidth]{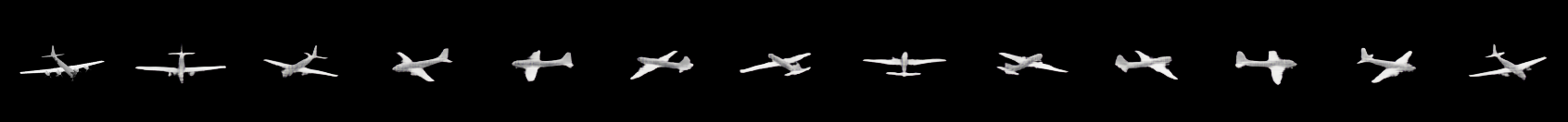}
    \includegraphics[width=\textwidth]{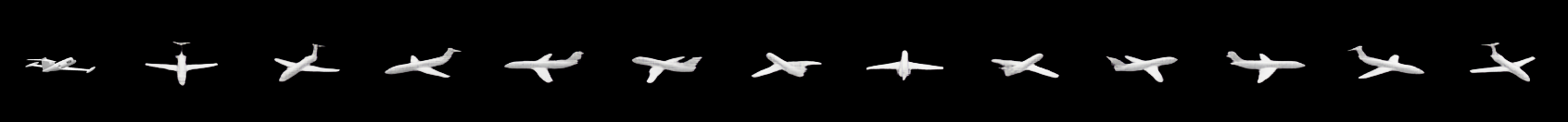}
    \includegraphics[width=\textwidth]{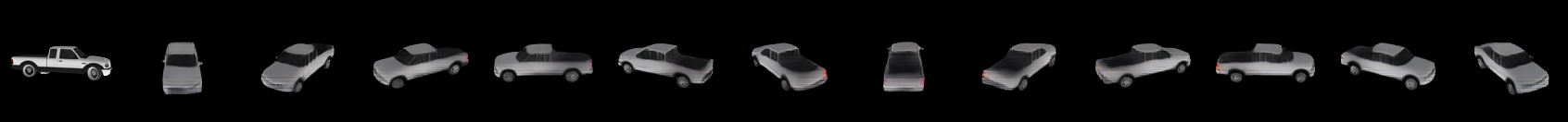}
    \includegraphics[width=\textwidth]{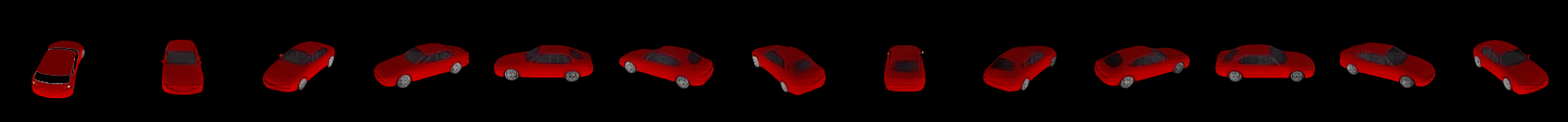}
    \includegraphics[width=\textwidth]{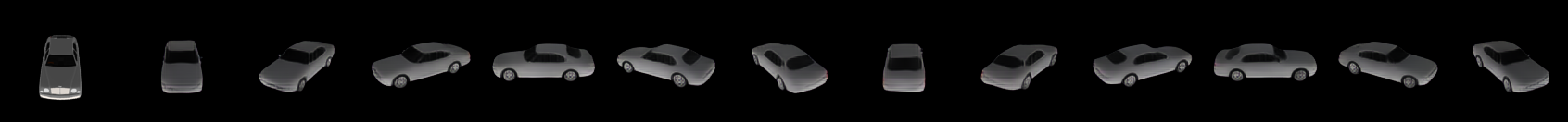}
    \includegraphics[width=\textwidth]{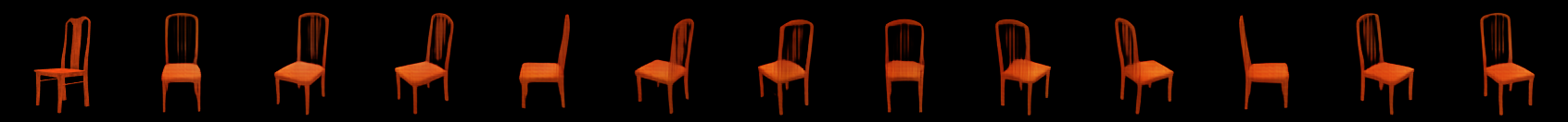}
    \includegraphics[width=\textwidth]{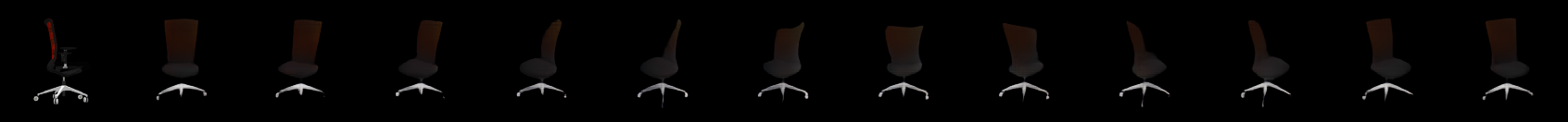}
    \includegraphics[width=\textwidth]{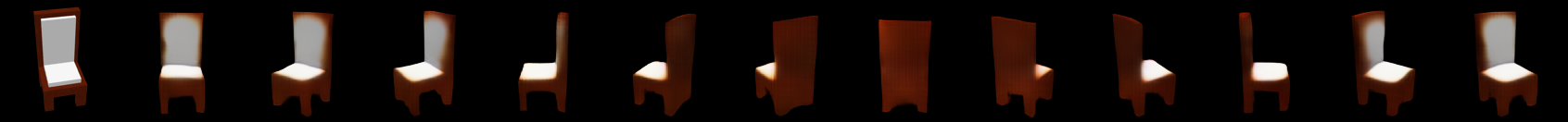}
    \caption{Novel view generation. Leftmost image provides object embedding}
    \label{fig:view_synthesis}
\end{figure}

We also demonstrate that our model is able to generate arbitrary views of an unseen object from a single image.  To do so, we feed an image to the encoder to obtain an embedding defining the identity of the object we want to generate views from. Then, we rotate it at the desired viewpoint, and decode it. Example results for all three categories are shown on figure~\ref{fig:view_synthesis}, with viewpoints picked every 30 degrees in azimuth from the origin. We can observe that prominent features defining the identity of the object - \eg global shape, texture - are preserved, and the viewpoints are correctly spaced. Of course, as with any other method doing view synthesis, errors occur as the model has to fill in the parts of the object that are self-occluded. This results on the loss of finer details, like spindles between chair legs. However, the correctness of the viewpoints means those pictures could be used to further refine predictions.

%% file: conclusion.tex

We introduced an approach able to leverage unlabeled data in order to efficiently learn viewpoint estimation with minimal supervision. By learning a geometry-aware representation of objects, our framework can use self-supervision as a proxy, retaining reasonable performances when viewpoint supervision is scarce. Our experiments show that we outperform simple supervised approaches at equal supervision, as well as other state of the art semi-supervised methods. Our method can as well produce new views of objects, allowing it to be used as a generative model. While it does show its limitation in cased of extreme supervision, we hope that proper regularization dealing with symmetries can solve the issue and allow for completely unsupervised pose estimation.